\documentclass[letterpaper]{article}
\usepackage[preprint]{aaai2027}
\usepackage[hyphens]{url}
\usepackage{graphicx}
\usepackage{natbib}
\usepackage{caption}
\usepackage[utf8]{inputenc}
\usepackage{booktabs}
\usepackage{amsfonts}
\usepackage{amsmath}
\usepackage{multirow}
\usepackage{pifont}
\newcommand{\cmark}{\ding{51}}%
\newcommand{\xmark}{\ding{55}}%
\usepackage{xspace}
\newcommand{\methodname}{\mbox{\textit{VADER}}\xspace}

\title{VADER: Adaptive Debiasing for Hallucination Mitigation in Video Large Language Models}
\author{%
  Dong Xing\textsuperscript{1,2} \quad
  Jiaxin Chen \quad
  Hang Yang\textsuperscript{1,2} \quad
  Peixun Liu\textsuperscript{1,2}\footnotemark[1] \\
  Qiushi Yang\textsuperscript{3}\thanks{Corresponding authors.} \quad
  Yuqing Wang\textsuperscript{1,2}\footnotemark[1]
}
\affiliations{%
  \textsuperscript{1}Changchun Institute of Optics, Fine Mechanics and Physics, Chinese Academy of Sciences\\
  Changchun, Jilin 130033, China\\
  \textsuperscript{2}University of Chinese Academy of Sciences, Beijing 100049, China\\
  \textsuperscript{3}City University of Hong Kong, Hong Kong, China
}

\begin{document}

\maketitle

\begin{abstract}
Large vision-language models (LVLMs) have demonstrated strong performance in open-ended video understanding, yet they remain prone to fluent responses unsupported by video evidence. Existing training-free methods typically apply a globally fixed visual intervention or construct a contrastive branch through input perturbation. The former cannot accommodate video-dependent fusion paths, while the latter can be compensated by cross-frame redundancy. We therefore propose Video-Adaptive Debiasing via Evidence Reweighting (\methodname), a training-free framework with two complementary modules. Visual Focus Reallocation (VFR) automatically instantiates an intervention policy for each video--question input: it diagnoses layer-wise visual-to-text evidence flow, determines where to intervene, and derives how strongly to reallocate pre-softmax attention from system-token to video-token blocks. Selective Evidence Erasure (SEE) independently masks high-importance visual tokens in every frame, constructing a prior-biased branch that is difficult to compensate through neighboring frames. Contrastive decoding then down-weights predictions that remain confident after selective evidence erasure. Across multiple VideoLLMs, \methodname yields substantial improvements on event-level grounding and temporal consistency; on LLaVA-Video-7B, it reaches 72.60\% accuracy on EventHallusion.
\end{abstract}

\section{Introduction}

Video vision-language models~\cite{llava,qwenvl,gpt4,gemini} (LVLMs) are driving video understanding beyond classification and retrieval toward open-ended question answering, event summarization, and long-horizon temporal reasoning~\cite{zhang2024llavanextvideo,li2023videochat,lin2024video,maaz2024video,ding2023hilm,ding2024holistic}. Nevertheless, these models remain susceptible to hallucinations during generation: their responses may be linguistically fluent and plausible while lacking support from the actual video evidence, with errors occurring particularly often in action recognition, event ordering, causal relationships, and temporal consistency. Unlike static images~\cite{vcd,SID,Vasparse,yin2024woodpecker,lee2024volcano,zhou2023analyzing,marine}, videos contain abundant redundant frames~\cite{AKS,lei2023revealing}, sparse key evidence, and complex cross-frame dependencies~\cite{song2024moviechat}, making models more likely to fall back on shortcut reasoning driven by language priors. Improving video grounding during decoding without retraining the model has therefore become a key problem for deploying video LVLMs.

Existing hallucination mitigation methods can be broadly categorized into training-based approaches and training-free decoding interventions. Training-based methods rely on high-quality visual instruction data, preference optimization~\cite{tpo,rrpo,ding2025pami}, reinforcement learning~\cite{arrowrl}, or video-specific post-training objectives~\cite{ding2025pami}. While these methods often provide stable gains, they require additional supervision, reward models, or costly retraining. Training-free methods instead directly modify model behavior during decoding and are thus better suited as plug-and-play deployment strategies. Representative approaches include enhancing visual attention or visual features using fixed rules~\cite{ICD}, as well as constructing contrastive branches through input noise, local masking, or frame dropping~\cite{eventhallusion}. Videos pose a distinct challenge: consecutive frames are highly redundant yet temporally dependent. Evidence removed by local perturbations can often be recovered from neighboring frames, whereas coarse frame removal may disrupt the dependencies required for temporally consistent event reasoning. Meanwhile, video evidence must compete for attention against long system prompts, instruction templates, and contextual tokens.

A closer examination of existing training-free methods reveals two video-specific bottlenecks. The first is \textit{static modality intervention}: a globally prescribed layer range and strength cannot accommodate the video-dependent depths at which visual evidence competes with system prompts, historical text, and massive video-token sequences. The second is \textit{inadequate contrastive evidence}: coarse input perturbations~\cite{eventhallusion,wu2025season} either leave evidence recoverable across redundant frames or damage temporal dependencies. Effective video hallucination mitigation should instead determine an intervention policy from each input and erase key evidence across the temporal axis while preserving low-response temporal context.

To this end, we propose Video-Adaptive Debiasing via Evidence Reweighting (\methodname), a training-free hallucination mitigation framework for video LVLMs. VFR treats visual enhancement as an automated, video-conditioned policy rather than a fixed configuration: diagnostic visual-to-text flow jointly determines the intervention layers and response strength for the current input. At these layers, VFR reallocates pre-softmax attention between system-token and video-token blocks. SEE then retains low-importance tokens and masks high-importance tokens inside the decoder on a per-frame basis, constructing a prior-biased branch that is difficult to recover through temporal redundancy. Finally, \methodname contrasts the grounded and prior-biased logits to suppress weakly grounded predictions. The main contributions are:
\begin{itemize}
    \item We propose \methodname, a training-free framework for mitigating hallucinations in video LVLMs through a dual-branch design that calibrates evidence grounding and suppresses prior-biased predictions.
    \item VFR formulates visual enhancement as a video-conditioned intervention policy that automatically determines both where and how strongly to recalibrate system--video attention.
    \item We introduce SEE, which performs decoder-internal, per-frame erasure of high-importance visual tokens to construct a cleaner prior-biased branch for contrastive suppression.
    \item \methodname achieves strong gains on event-level hallucination benchmarks, reaching 72.60\% on EventHallusion with LLaVA-Video-7B and surpassing TPO and RRPO.

\end{itemize}

\section{Related Work}
\label{sec:related_work}
\subsection{Video Large Language Models}
Multimodal Large Language Models (MLLMs) couple language backbones with visual encoders for cross-modal perception and reasoning~\cite{llava,qwenvl,gpt4,gemini}. VideoLLMs extend this paradigm by injecting spatiotemporal visual tokens into LLM decoders, enabling open-ended understanding of dynamic scenes~\cite{llavavideo,qwen25vl,lin2024video}. Representative models such as Video-LLaVA~\cite{lin2024video}, VideoChatGPT~\cite{maaz2024video}, Valley~\cite{wu2025valley2}, and LLaVA-Video achieve strong results on video QA and general video understanding benchmarks~\cite{videomme,mvbench,tempcompass,tvbench}. However, benchmark accuracy does not guarantee faithful grounding: recent studies show that VideoLLMs still hallucinate under event ordering, motion reasoning, causal inference, and long-range temporal dependencies~\cite{videohallucer,eventhallusion,vidhalluc}. This motivates parameter-free mechanisms that improve video-grounded generation during decoding.

\subsection{Hallucination Mitigation in Multimodal Models}
Existing mitigation methods can be broadly divided into training-based and training-free approaches. Training-based methods enhance cross-modal alignment using instruction tuning~\cite{lrvinstruction,liumitigating,yu2024hallucidoctor}, human feedback~\cite{rlhfv}, preference optimization~\cite{tpo,rrpo,ding2025pami}, reinforcement learning~\cite{arrowrl}, or video-specific representation/activation design~\cite{taae,bae2025mash}. Despite their effectiveness, they often require additional supervision, reward modeling, or parameter updates. Training-free methods instead modify decoding through contrastive prediction~\cite{vcd,eventhallusion}, auxiliary visual guidance~\cite{marine}, or attention recalibration~\cite{avisc,opera,ma2024vista}. Saliency can identify visually responsive components, but a globally prescribed intervention configuration remains insensitive to video-dependent fusion paths. VFR instead closes the loop from diagnosis to intervention: every video--question input receives its own layer set and response strength. SEE complements this automated policy by constructing a decoder-internal, frame-wise prior branch that resists temporal compensation.

\section{Method}

\begin{figure*}[t]
    \centering
    \includegraphics[width=\textwidth]{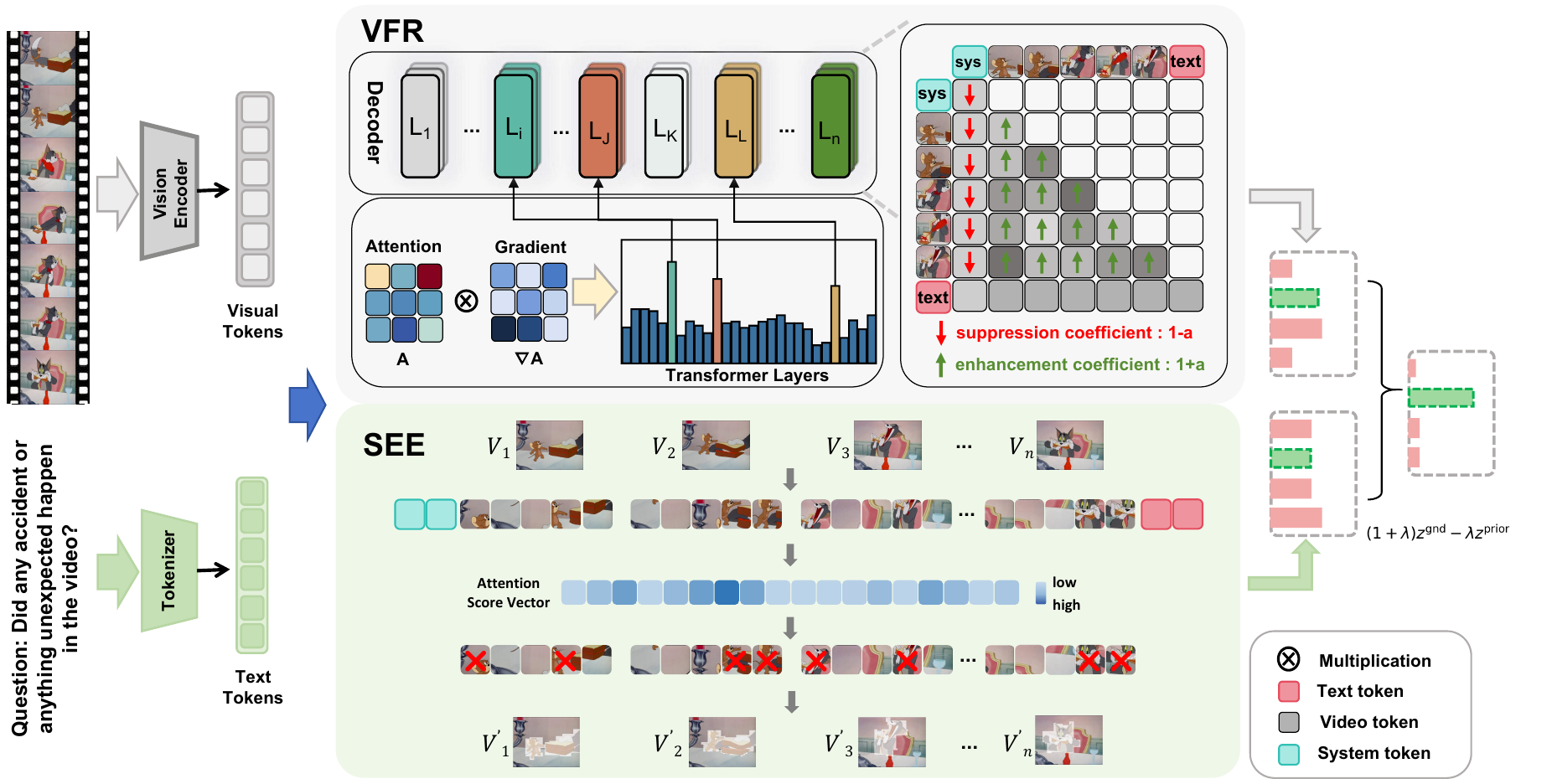}
    \caption{Overview of the \methodname framework. VFR automatically derives a video-conditioned intervention policy---the layer locations and response strength---and reallocates attention toward video evidence. SEE suppresses high-response visual tokens independently in every frame to expose prior-biased predictions. Joint temporal contrastive suppression contrasts the two branches.}
    \label{fig:framework}
\end{figure*}

\subsection{Preliminaries}

\paragraph{Contrastive Decoding.} Given a long video $V=\{f_1,\dots,f_T\}$ and a textual query $q$, a VideoLLM performs autoregressive decoding conditioned on the encoded visual and textual context. At each decoding step, let $z^{+}$ denote the logits from the main branch and $z^{-}$ denote the logits from a degraded branch. Contrastive decoding is typically formulated as

\begin{equation}
z^{\text{cd}} = (1+\lambda)z^{+} - \lambda z^{-}, \qquad \lambda \ge 0,
\end{equation}

which suppresses candidate tokens that remain highly confident under degraded conditions. \methodname follows this basic formulation, but its focus is not on modifying the contrastive equation itself. Instead, it aims to construct more reliable $z^{+}$ and $z^{-}$ for long videos: the former should make fuller use of video evidence, while the latter should more closely reflect language-prior-driven predictions.

\paragraph{Attention Notation.} For the $h$-th attention head in the $l$-th decoder layer, let the query, key, and value be denoted by $Q^{(l,h)},K^{(l,h)},V^{(l,h)}$. Standard attention first computes the pre-softmax score

\begin{equation}
S^{(l,h)}_{ij}=Q^{(l,h)}_i K^{(l,h)\top}_j/\sqrt{d},
\end{equation}

then obtains $A^{(l,h)}_{ij}=\mathrm{softmax}_j(S^{(l,h)}_{ij})$ and aggregates the values. In the following, VFR reweighting directly acts on specified row-column blocks of the pre-softmax attention-score matrix $S^{(l,h)}$, while SEE erasure can be written as an additive mask applied to $S^{(l,h)}$. Setting certain positions to $-\infty$ indicates that the corresponding token positions are masked before softmax. We use $\mathcal{T}_{\text{sys}}$, $\mathcal{T}_{\text{vid}}$, $\mathcal{T}_{\text{txt}}$, and $\mathcal{T}_{\text{dec}}$ to denote system tokens, video tokens, textual context tokens, and the decoder-side positions actually intervened by VFR, respectively.

\paragraph{Overall Framework.} With the above notation, \methodname modifies the autoregressive decoding process without updating any LVLM parameters. The input video and query are first encoded into video tokens, system tokens, and textual context tokens. The same decoding state is then fed into two decoding branches: VFR produces the evidence-grounded branch logits $z^{\text{gnd}}$, while SEE produces the prior-biased branch logits $z^{\text{prior}}$. Finally, Joint Temporal Contrastive Suppression fuses the two sets of logits into $\bar{z}$ and returns them to standard next-token decoding. Sections 3.2--3.4 detail the two branches and their logits fusion.

\subsection{Visual Focus Reallocation}

\noindent\textbf{Visual--text Saliency Flow and Video-conditioned Policy.} Cross-modal evidence propagation is not uniformly distributed across decoder layers. Different video--question inputs can activate distinct, potentially non-contiguous fusion paths because their scenes, temporal structure, and relevant evidence differ. A single global layer configuration may therefore perturb layers unrelated to the current decision. VFR instead diagnoses layer-wise visual--text flow for each video--question input and automatically determines where evidence reallocation should occur.

Specifically, we use gradient--attention saliency to measure the contribution of each layer's attention to the current generation decision. Before generation, VFR performs a diagnostic forward pass using only the video--question input. Let $\hat{y}_1=\arg\max_y p_\theta(y\mid V,q)$ denote the unmodified model's own initial next-token prediction. Treating $\hat{y}_1$ as a detached target, we define $\mathcal{L}_{\text{diag}}=-\log p_\theta(\hat{y}_1\mid V,q)$. This objective is used only to obtain attention gradients for attribution and does not update any model parameters; no ground-truth answer, candidate label, or external supervision is accessed at test time. The resulting layer set and response strength are held fixed during subsequent autoregressive decoding. Let $A^{(l,h)}$ denote the post-softmax attention matrix of the $h$-th attention head in the $l$-th layer. The attention saliency tensor of layer $l$ is defined as
\begin{equation}
I^{(l)}_{ij} \;=\; \frac{1}{H}\sum_{h=1}^{H}
\bigl|\,A^{(l,h)}_{ij}\odot
\partial\mathcal{L}_{\text{diag}}/\partial A^{(l,h)}_{ij}\,\bigr|.
\end{equation}
The product term combines attention mass with its first-order sensitivity to the diagnostic loss, thereby highlighting token interactions that are both actually attended to by the model and influential for the current output distribution. Compared with simply measuring attention mass, this quantity is better suited for identifying cross-modal pathways that genuinely affect the current prediction.

We then read two types of visual-related flows from $I^{(l)}$: the intra-visual flow within video positions, and the cross-modal flow between textual positions and video positions. The former is mainly used to diagnose whether visual representations are stable; VFR layer selection relies only on the latter, because it directly characterizes how strongly the current decoding state extracts evidence from video tokens. Given the textual token set $\mathcal{T}_{\text{txt}}$ and the video token set $\mathcal{T}_{\text{vid}}$, the cross-modal saliency flow of layer $l$ is defined as
\begin{equation}
\phi_l \;=\; \frac{1}{|\mathcal{T}_{\text{txt}}|\,|\mathcal{T}_{\text{vid}}|}
\sum_{i\in \mathcal{T}_{\text{txt}}}
\sum_{j\in \mathcal{T}_{\text{vid}}} I^{(l)}_{ij}.
\end{equation}
This quantity corresponds to the visual-to-text flow component in our implementation. To eliminate scale differences in saliency values across samples, we perform min--max normalization over all decoder layers within each sample:
\begin{equation}
\bar{\phi}_l
=\frac{\phi_l-\min_m \phi_m}{\max_m \phi_m-\min_m \phi_m+\epsilon}.
\end{equation}
We then select the Top-$K_l$ layers with the highest normalized cross-modal flow as the reweighting layers:
\begin{equation}
\mathcal{L}_{\text{sel}} \;=\;
\operatorname{TopK}_{\text{largest}}
\bigl(\{\bar{\phi}_l\}_{l=1}^{L},\; K_l\bigr).
\end{equation}
This diagnostic-driven selection avoids a prescribed depth range and supports non-contiguous layers along the current video's evidence path. We use $K_l=5$ by default; sensitivity results are reported in the supplement.

\noindent\textbf{Evidence Strength and Scaling Amplitude.} After determining $\mathcal{L}_{\text{sel}}$, VFR estimates the evidence strength of the current sample using the normalized cross-modal flow of the selected layers, i.e., $x=\sum_{l\in\mathcal{L}_{\text{sel}}}\bar{\phi}_l$. A larger $x$ indicates that the current sample contains a clearer text--video evidence channel, and therefore stronger visual enhancement and system suppression can be applied. We map $x$ to a reweighting amplitude with sublinear growth:
\begin{equation}
a \;=\; \log\!\bigl(1 + \beta\,x\bigr).
\end{equation}
Here, $\beta$ maps saliency flow $x$ to response amplitude $a$. We use the logarithmic mapping in Eq.~(7), followed by the clipping described below, to prevent excessive intervention; $\beta=0.05$ by default.

Together, $\mathcal{L}_{\mathrm{sel}}$ and $a$ define the video-conditioned policy $\Pi(V,q)=(\mathcal{L}_{\mathrm{sel}},a)$: the former determines \emph{where} to intervene, while the latter determines \emph{how strongly}. Thus, VFR is not a fixed layer-ranking heuristic, but a closed-loop diagnosis--decision--intervention procedure instantiated anew for each input.

It is worth noting that $a$ does not act on key states, nor does it modify the query, key, or value projections themselves. Instead, it is used to construct multiplicative reweighting factors on the pre-softmax attention scores. This design preserves the original computation of the $QK^\top/\sqrt{d}$ scores and only adjusts the relative allocation of the system-token and video-token column blocks in the pre-softmax attention-score matrix. In implementation, we apply numerical-stability clipping to $a$ to ensure that the suppression-side scaling factor remains non-negative. This clipping is a deterministic decoding operation and introduces no learnable parameters.

\noindent\textbf{Complementary Attention-Score Reweighting.} Given the selected layer set $\mathcal{L}_{\text{sel}}$ and the scaling amplitude $a$, VFR performs complementary recalibration of attention scores corresponding to different token regions in the selected layers. We construct
\begin{equation}
\gamma_{\text{sup}} = 1-a,\qquad
\gamma_{\text{enh}} = 1+a ,
\end{equation}
where $\gamma_{\text{sup}}$ is used to reduce the relative influence of the system-token score block, and $\gamma_{\text{enh}}$ is used to enhance the relative influence of the video-token score block. For any $l\in\mathcal{L}_{\text{sel}}$, VFR first computes
\begin{equation}
S^{(l,h)}_{ij}=Q^{(l,h)}_iK^{(l,h)\top}_j/\sqrt{d},
\end{equation}
following the original model, and then applies multiplicative reweighting to the target score regions before softmax:
\begin{equation}
\widetilde{S}^{(l,h)}_{ij}
=
\begin{cases}
\gamma_{\text{sup}}\,S^{(l,h)}_{ij},
& i\in\mathcal{T}_{\text{dec}},\; j\in\mathcal{T}_{\text{sys}},\\
\gamma_{\text{enh}}\,S^{(l,h)}_{ij},
& i\in\mathcal{T}_{\text{dec}},\; j\in\mathcal{T}_{\text{vid}},\\
S^{(l,h)}_{ij},
& \text{otherwise}.
\end{cases}
\end{equation}
Standard softmax and value aggregation are then performed:
\begin{equation}
\widetilde{A}^{(l,h)}_{ij}
= \mathrm{softmax}_j\!\left(\widetilde{S}^{(l,h)}_{ij}\right).
\end{equation}
This formulation emphasizes that VFR modifies only the system/video score blocks at the layers activated by $\Pi(V,q)$. The resulting grounded branch therefore reallocates attention toward video evidence without applying the same global intervention to every input.

\subsection{Selective Evidence Erasure}
\noindent\textbf{Per-frame Attention Masking.} Neighboring frames in long videos often contain substantial redundant information. Input-level masking or random frame dropping can therefore be easily compensated by other frames, making it difficult to form a sufficiently clean prior-biased branch. The core idea of SEE is to remove, inside the decoder and on a per-frame basis, the visual tokens on which the current prediction most depends. This allows the negative branch to retain background and weak visual context while weakening the discriminative evidence that truly supports the answer.

Specifically, a VideoLLM flattens visual tokens from all video frames into a long sequence. SEE uses \texttt{frame\_ranges} recorded during visual encoding to recover the token interval $\mathcal{T}^{(t)}_{\text{frame}}$ for each frame, and estimates token importance at a fixed insertion layer $l^\star$. Let $q_{\mathrm{cur}}$ denote the current decoding position, and let $A^{(l^\star,h)}$ be the post-softmax attention matrix of the $h$-th attention head. The score of a visual token $v_j$ is defined as
\begin{equation}
\mathrm{Score}(v_j) \;=\;
\frac{1}{H}\sum_{h=1}^{H}
A^{(l^\star,h)}_{q_{\mathrm{cur}},j},
\qquad v_j\in\mathcal{T}_{\text{vid}} .
\end{equation}
A higher score indicates that the token is more likely to provide discriminative visual evidence for the current decoding step. Unlike global token selection, SEE allocates the retention budget to each frame and retains the $k_t$ lowest-scoring tokens within each frame:
\begin{equation}
\mathcal{R}^{(t)}
=
\operatorname{BottomK}_{k_t}
\bigl(\mathcal{T}^{(t)}_{\text{frame}};\mathrm{Score}\bigr).
\end{equation}
Here $k_t$ is determined by the retention ratio $\rho$, i.e., $k_t=\lfloor \rho |\mathcal{T}^{(t)}_{\text{frame}}|\rfloor$. This per-frame budget prevents erasure from being concentrated on a small number of frames, thereby uniformly weakening discriminative visual evidence across the entire temporal axis. All video tokens that are not retained form the erasure set
\begin{equation}
\mathcal{M}
=
\mathcal{T}_{\text{vid}}
\setminus
\bigcup_{t=1}^{T}
\mathcal{R}^{(t)} .
\end{equation}
Finally, SEE applies an attention-score column mask to these tokens at the fixed insertion layer $l^\star$:
\begin{equation}
\widehat{S}^{(l^\star,h)}_{ij}
=
S^{(l^\star,h)}_{ij}
+
\begin{cases}
-\infty, & j\in\mathcal{M}, \\
0, & \text{otherwise}.
\end{cases}
\end{equation}
In this way, each frame still preserves a subset of low-response visual context, while high-response foreground evidence is removed. Unlike input-level perturbations, SEE does not modify the output of the visual encoder, but instead constructs a negative branch inside decoder attention that is more biased toward background and language priors. The resulting logits are denoted as $z^{\text{prior}}$.


\noindent\textbf{Insertion Depth.}
The SEE insertion layer $l^\star$ determines when visual evidence is erased inside the decoder. An overly early insertion yields unstable visual saliency before cross-modal representations are sufficiently formed, whereas an overly late insertion leaves little room to create a distinguishable prior-biased branch. We therefore set $l^\star$ as a fixed configuration selected by layer-wise diagnostics: we compute query-conditioned spatial saliency maps and choose the earliest layer where visual responses become stable and concentrated while sufficient decoding depth remains. This places SEE after visual evidence becomes localizable but before it is fully absorbed into the final logits.

\subsection{Joint Temporal Contrastive Suppression}

VFR and SEE respectively produce the evidence-grounded branch logits $z^{\text{gnd}}$ and the prior-biased branch logits $z^{\text{prior}}$. To exploit both branches during decoding, \methodname fuses them via contrastive suppression:

\begin{equation}
\bar{z} \;=\; (1+\lambda)\, z^{\text{gnd}} - \lambda\, z^{\text{prior}},
\end{equation}

where $\lambda\ge 0$ is the contrastive-strength hyperparameter and is independent of the attention-score reweighting amplitude $a$ in Section 3.2. The mechanism of this formulation is straightforward: if a candidate token receives high scores in both the grounded branch and the prior-biased branch, its high score is more likely to originate from a language prior rather than video evidence, and $\bar{z}$ will suppress it accordingly. If a candidate token is mainly contributed by the grounded branch and is not dominant in the prior-biased branch, its relative score is preserved or amplified in $\bar{z}$.

\section{Experiments}
\subsection{Experimental Setup}
\noindent\textbf{Benchmark Datasets.} We evaluate \methodname on three video hallucination benchmarks, \textbf{VidHalluc}~\cite{vidhalluc}, \textbf{VideoHallucer}~\cite{videohallucer}, and \textbf{EventHallusion}~\cite{eventhallusion}, which assess object, action, event, temporal, and semantic grounding errors. To examine whether hallucination mitigation preserves general video understanding, we further report results on \textbf{MVBench}~\cite{mvbench}, \textbf{VideoMME}~\cite{videomme}, and \textbf{TVBench}~\cite{tvbench}. Experiments are conducted on three open-source VideoLLM backbones, \textbf{LLaVA-OneVision-7B}~\cite{llavaov}, \textbf{LLaVA-NeXT-Video-7B}~\cite{llavavideo}, and \textbf{Qwen3-VL-8B}~\cite{qwen3vl}, covering different multimodal architectures and visual tokenization schemes. We compare against the original decoding strategy, representative hallucination mitigation methods including DINO-HEAL~\cite{vidhalluc} and TCD~\cite{eventhallusion}, and ablated variants of \methodname. Unless otherwise specified, all models use publicly available pretrained weights without additional fine-tuning.

\noindent\textbf{Implementation and Evaluation Details.} All experiments follow a \textbf{training-free} protocol: \methodname is inserted into the original generation pipeline of each backbone, with the visual encoder, projection layer, and language decoder kept frozen. Each video is tokenized by the pretrained visual encoder and decoded together with the textual context by the frozen language model. We select a single global configuration on EventHallusion with LLaVA-Video-7B and apply it unchanged to all other benchmarks and backbones, without target-specific retuning. Specifically, $\beta=0.05$, $\lambda=1.5$, $K_l=5$, $l^\star=8$, and $\rho=0.3$. Deterministic decoding is adopted in the prior-biased branch for stability.

\subsection{Experimental Results}
\noindent\textbf{Quantitative Evaluation.} Table \ref{tab:halbanchmark} summarizes the quantitative results of \methodname on the three video hallucination benchmarks. Overall, \methodname's gains are most prominent on EventHallusion and VideoHallucer, which emphasize event-level grounding and temporal consistency, while its performance on VidHalluc exhibits some model dependence. For EventHallusion, \methodname improves the accuracy of Qwen3-VL-8B, LLaVA-OV-7B, and LLaVA-Video-7B from \textbf{65.00\%}, \textbf{60.15\%}, and \textbf{63.57\%} to \textbf{71.00\%}, \textbf{68.84\%}, and \textbf{72.60\%}, corresponding to gains of \textbf{+6.00}, \textbf{+8.69}, and \textbf{+9.03} points, respectively. On the average results of VideoHallucer, \methodname brings improvements of \textbf{+3.90}, \textbf{+10.30}, and \textbf{+7.29} points across the three backbones. For LLaVA-OV-7B, \methodname obtains an EventHallusion result close to the strongest training-free baseline while achieving higher average scores on VidHalluc and VideoHallucer. For Qwen3-VL-8B, \methodname improves the overall VidHalluc average from 60.37 to 62.60 but yields clear gains on VideoHallucer and EventHallusion. Notably, although \methodname requires no additional training, on LLaVA-Video-7B it still achieves \textbf{72.60\%} accuracy on EventHallusion, surpassing the training-based methods TPO (\textbf{63.33\%}) and RRPO (\textbf{67.97\%}).

\begin{table*}[t]
    \centering
    \small
    \setlength{\tabcolsep}{1.3pt}
    \begin{tabular}{lccccccccccccc}
    \toprule
    \multirow{2}{*}{\textbf{Models}} & \multirow{2}{*}{\textbf{\shortstack[c]{Training-\\free}}}
    & \multicolumn{5}{c}{\textbf{VidHalluc}}
    & \multicolumn{6}{c}{\textbf{VideoHallucer}}
    & \multicolumn{1}{c}{\textbf{EventHallusion}} \\
    \cmidrule(lr){3-7} \cmidrule(lr){8-13} \cmidrule(lr){14-14}
    & & BQA & MCQ & STH & TSH & AVG
    & ORH & TPH & SDH & EFH & ENFH & AVG
    & AVG \\
    
    \midrule
    Qwen3-VL-8B~\cite{qwen3vl} & \textbf{-} & 54.79 & \textbf{87.53} & 41.52 & \textbf{57.67} & 60.37 & \textbf{78.52} & 75.00 & 79.40 & 61.48 & \textbf{94.50} & 77.90 & 65.00 \\
    +TCD~\cite{eventhallusion} & \cmark & 55.98 & 86.74 & 45.83 & 57.56 & 61.53 & 77.84 & 81.50 & 79.50 & 63.92 & 94.32 & 79.52 & 67.38 \\
    +DINO-HEAL~\cite{vidhalluc} & \cmark & 56.61 & 86.21 & 47.62 & 57.59 & 62.01 & 77.26 & 84.00 & 79.80 & 65.84 & 94.18 & 80.26 & 68.95 \\
    \textbf{+\methodname (Ours)} & \cmark & \textbf{57.17} & 85.63 & \textbf{50.10} & 57.50 & \textbf{62.60} & 76.00 & \textbf{91.00} & \textbf{80.00} & \textbf{68.00} & 94.00 & \textbf{81.80} & \textbf{71.00} \\
    \midrule
    
    LLaVA-OV-7B~\cite{llavaov} & \textbf{-} & 74.36 & 90.27 & \textbf{63.65} & 53.00 & 70.32 & 56.50 & 52.50 & \textbf{56.50} & 15.00 & 51.50 & 46.40 & 60.15 \\
    +TCD~\cite{eventhallusion} & \cmark & 72.44 & 90.06 & 58.57 & 64.33 & 71.35 & 59.50 & 53.50 & 56.00 & 17.50 & 54.00 & 48.10 & 68.46 \\
    +DINO-HEAL~\cite{vidhalluc} & \cmark & 74.29 & 90.36 & 63.18 & 53.00 & 70.21 & 57.00 & 53.50 & \textbf{56.50} & 15.50 & 52.00 & 46.90 & 60.15 \\
    +SEASON~\cite{wu2025season} & \cmark & 73.15 & 90.51 & 60.29 & 77.50 & 75.36 & \textbf{63.00} & 55.50 & \textbf{56.50} & 19.50 & 48.00 & 48.50 & \textbf{69.19} \\
    \textbf{+\methodname (Ours)} & \cmark & \textbf{77.36} & \textbf{90.95} & 62.76 & \textbf{80.83} & \textbf{77.98} & 54.50 & \textbf{83.52} & 51.00 & \textbf{28.00} & \textbf{66.50} & \textbf{56.70} & 68.84 \\
    \midrule
    
    LLaVA-Video-7B~\cite{llavavideo} & \textbf{-} & 75.02 & 90.76 & 51.23 & 38.00 & 63.75 & 60.00 & 61.50 & 66.50 & 16.50 & 52.50 & 51.40 & 63.57 \\
    +TCD~\cite{eventhallusion} & \cmark & 73.50 & 90.41 & 49.64 & 46.00 & 64.89 & 58.00 & 61.00 & 65.50 & 14.50 & 51.50 & 50.10 & 64.30 \\
    +DINO-HEAL~\cite{vidhalluc} & \cmark & 75.27 & 90.81 & 51.51 & 37.67 & 63.81 & 59.50 & 61.00 & 66.50 & 17.00 & 52.50 & 51.30 & 64.30 \\
    +TPO~\cite{tpo} & \xmark & 74.85 & 90.69 & 49.62 & 42.50 & 64.42 & 60.00 & 59.50 & \textbf{68.50} & 16.00 & 52.50 & 51.30 & 63.33 \\
    +RRPO~\cite{rrpo} & \xmark & \textbf{76.80} & 91.23 & 49.83 & 37.67 & 63.88 & 59.00 & 58.00 & 67.00 & \textbf{21.00} & 51.50 & 51.30 & 67.97 \\
    +SEASON~\cite{wu2025season} & \cmark & 74.71 & 90.95 & 49.86 & 50.33 & 66.46 & 60.50 & 62.00 & 68.00 & 18.00 & 48.50 & 51.40 & 66.99 \\
    \textbf{+\methodname (Ours)} & \cmark & 71.45 & \textbf{91.32} & \textbf{55.35} & \textbf{64.17} & \textbf{70.57} & \textbf{64.50} & \textbf{82.95} & 66.50 & 17.50 & \textbf{62.00} & \textbf{58.69} & \textbf{72.60} \\
    \bottomrule
    \end{tabular}
    \caption{Evaluation on video hallucination benchmarks. Results on VidHalluc, VideoHallucer, and EventHallusion with different VideoLLM backbones. All numbers are percentages. All \methodname results use the same globally fixed configuration.}
    \label{tab:halbanchmark}
\end{table*}

\noindent\textbf{Preservation of general video understanding.} We further test whether improved faithfulness comes at the cost of standard video understanding across diverse evaluation settings. As shown in Table~\ref{tab:general_video}, \methodname increases the average score by \textbf{+0.6} and \textbf{+1.5} points on LLaVA-OV-7B and LLaVA-Video-7B, respectively, while competing interventions are mostly neutral or slightly negative.  These results indicate that \methodname improves video-grounded faithfulness without sacrificing broad perception, temporal understanding, and reasoning ability.

\begin{table}[t]
    \centering
    \small
    \setlength{\tabcolsep}{1.0pt}
    \begin{tabular}{lccccc}
    \toprule
    \textbf{Models} & \textbf{MV} & \textbf{MME} & \textbf{TV} & \textbf{AVG} & \textbf{$\Delta$} \\
    \midrule
    LLaVA-OV-7B & 54.4 & 50.6 & 42.8 & 49.3 & -- \\
    +TCD & 54.3 & 50.4 & 42.9 & 49.2 & $-0.1$ \\
    +DINO-HEAL & 54.4 & 50.6 & 42.9 & 49.3 & $0.0$ \\
    \textbf{+\methodname~(Ours)} & \textbf{55.2} & \textbf{51.3} & \textbf{43.1} & \textbf{49.9} & \textbf{$+0.6$} \\
    \midrule
    LLaVA-Video-7B & 58.4 & 53.0 & 45.2 & 52.2 & -- \\
    +TCD & 57.5 & 53.0 & 45.2 & 51.9 & $-0.3$ \\
    +DINO-HEAL & 58.3 & 53.2 & 45.5 & 52.3 & $+0.1$ \\
    +TPO & 58.2 & 53.0 & 45.1 & 52.1 & $-0.1$ \\
    +RRPO & 58.1 & 53.8 & 45.2 & 52.4 & $+0.2$ \\
    \textbf{+\methodname~(Ours)} & \textbf{59.8} & \textbf{54.5} & \textbf{46.9} & \textbf{53.7} & \textbf{$+1.5$} \\
    \bottomrule
    \end{tabular}
    \caption{General video understanding. $\Delta$ denotes the change over each backbone. MV, MME, and TV denote MVBench (multiple choice), VideoMME (without subtitles), and TVBench (multiple choice), respectively.}
    \label{tab:general_video}
\end{table}

\subsection{VFR as a Video-Conditioned Policy}
\begin{figure}[t]
    \centering
    \includegraphics[width=\columnwidth]{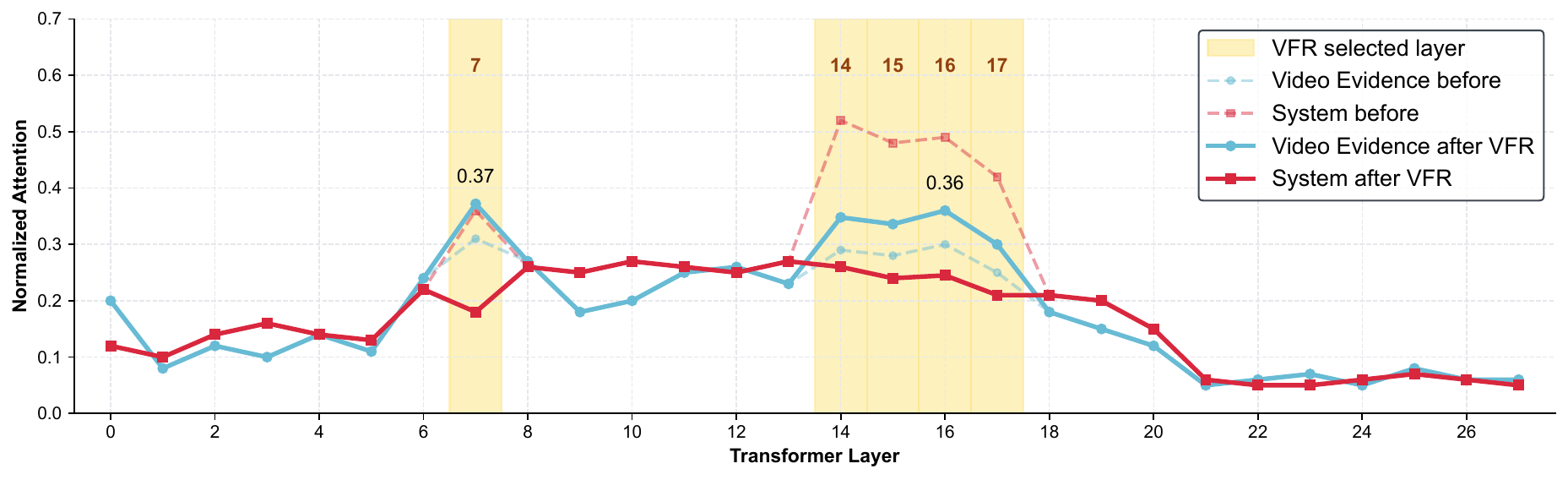}
    \caption{Attention rebalancing by fixed-layer VFR.}
    \label{fig:vsar_rebalance_curve}
\end{figure}

To isolate reweighting from policy construction, we apply VFR at fixed layers $7,14,15,16,17$ for every video, disabling adaptive selection. Table~\ref{tab:vfr_fixed_layer} shows gains from $63.75/51.40/63.57$ to $67.32/54.63/67.92$ on VidHalluc, VideoHallucer, and EventHallusion. In contrast, the VFR-only ablation in Table~\ref{tab:ablation_main} retains per-video selection and adaptive strength but disables SEE; its stronger results demonstrate the value of automatic policy construction. Figure~\ref{fig:vsar_rebalance_curve} confirms its system-to-video attention shift.

\begin{table}[t]
    \centering
    \small
    \setlength{\tabcolsep}{4.0pt}
    \begin{tabular}{lccc}
    \toprule
    \textbf{Setting} & \textbf{Vid.} & \textbf{VHall.} & \textbf{EHall.} \\
    \midrule
    Base & 63.75 & 51.40 & 63.57 \\
    Fixed VFR & \textbf{67.32} & \textbf{54.63} & \textbf{67.92} \\
    \bottomrule
    \end{tabular}
    \caption{Fixed-layer VFR.}
    \label{tab:vfr_fixed_layer}
\end{table}

\subsection{Source-setting Calibration of SEE Insertion Depth}

\begin{figure*}[t]
    \centering
    \begin{minipage}[t]{0.5\textwidth}
    \centering
    \includegraphics[width=0.98\linewidth]{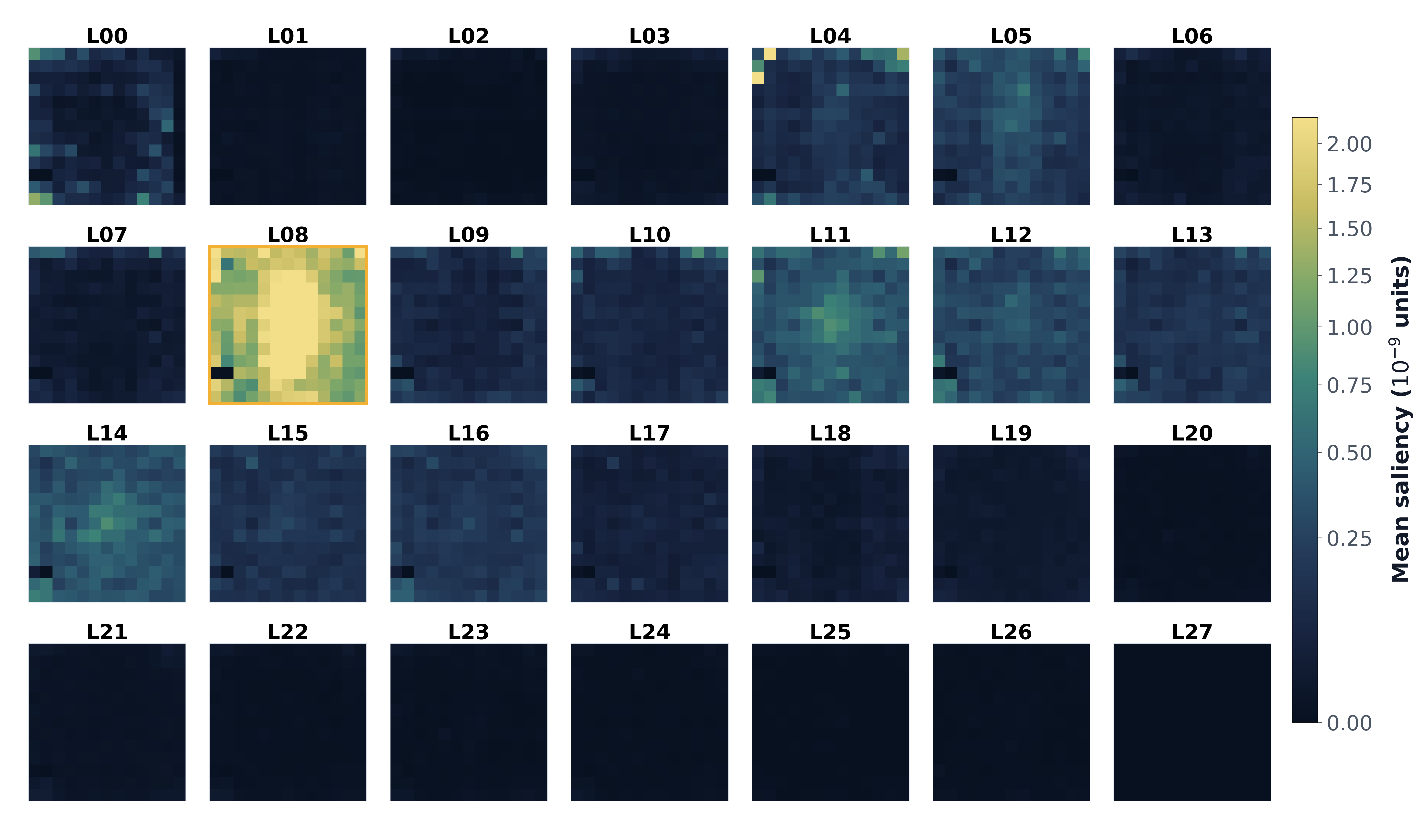}
    \end{minipage}
    \hfill
    \begin{minipage}[t]{0.48\textwidth}
    \centering
    \IfFileExists{tfd_layer.png}{
        \includegraphics[width=\linewidth]{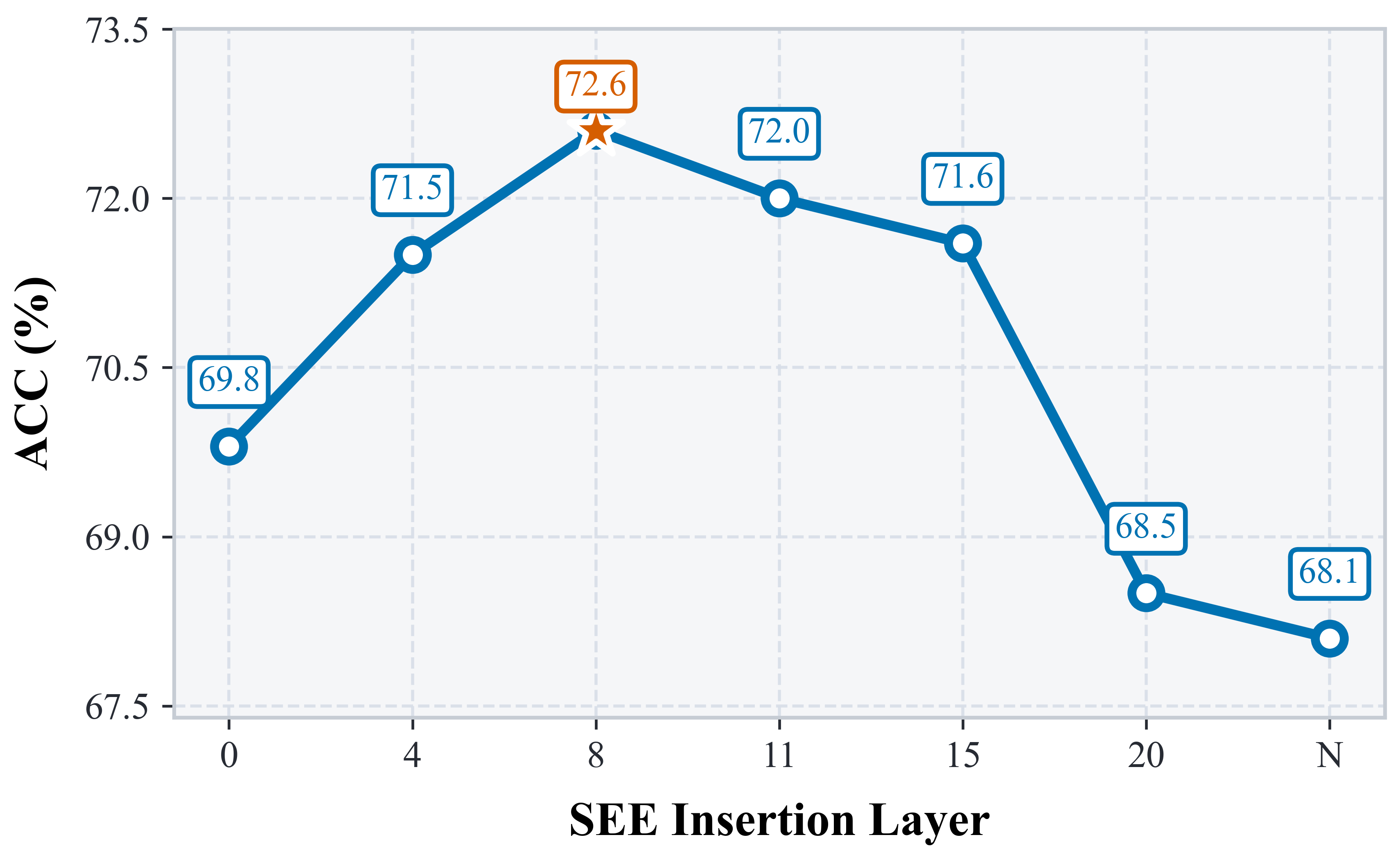}
    }{
        \fbox{\parbox{0.95\linewidth}{Pending figure generation: \texttt{tfd\_layer.png}.}}
    }
    \end{minipage}
    \vspace{-2mm}
    \caption{One-time calibration of the SEE insertion depth on the source setting. Left: query-conditioned saliency visualization of a representative example. Right: benchmark-level accuracy under different insertion layers on EventHallusion. The selected layer is fixed for all remaining evaluations.}
    \vspace{-1mm}
    \label{fig:tfd_layer}
\end{figure*}

We calibrate the SEE insertion depth using source-setting end-to-end evaluation and provide qualitative saliency visualization for interpretation. As shown in Figure~\ref{fig:tfd_layer}, the curve on the right compares the tested insertion depths: across the other settings, EventHallusion accuracy ranges from \textbf{68.1\%} to \textbf{72.0\%}, while layer 8 achieves the highest accuracy of \textbf{72.6\%} on LLaVA-Video-7B. The left heatmaps show query-conditioned saliency for one example. We select layer 8 once here and keep it fixed thereafter.

\subsection{Ablation Studies}
\noindent\textbf{Core-component ablation.} Table~\ref{tab:ablation_main} reports ablations on LLaVA-Video-7B. The VFR-only setting retains per-video selection and adaptive strength but disables SEE, unlike the fixed-layer variant in Table~\ref{tab:vfr_fixed_layer}. The complete \methodname improves EventHallusion, VideoHallucer, and VidHalluc from \textbf{63.57\%}/\textbf{51.40\%}/\textbf{63.75\%} to \textbf{72.60\%}/\textbf{58.69\%}/\textbf{70.57\%}, confirming the complementarity of VFR and SEE.

\begin{table}[t]
    \centering
    \small
    \setlength{\tabcolsep}{1.5pt}
    \begin{tabular}{lcc ccc}
    \toprule
    \multirow{2}{*}{\textbf{Method}} & \multirow{2}{*}{\textbf{VFR}} & \multirow{2}{*}{\textbf{SEE}} & \multicolumn{3}{c}{\textbf{Benchmark AVG (\%)}} \\
    \cmidrule(lr){4-6}
    & & & \textbf{EHall.} & \textbf{VHall.} & \textbf{VidHalluc} \\
    \midrule
    Baseline & \xmark & \xmark & 63.57 & 51.40 & 63.75 \\
    w/ VFR (only) & \cmark & \xmark & 68.45 & 56.73 & 69.46 \\
    w/ SEE (only) & \xmark & \cmark & 67.23 & 55.48 & 67.23 \\
    \textbf{\methodname (Ours)} & \cmark & \cmark & \textbf{72.60} & \textbf{58.69} & \textbf{70.57} \\
    \bottomrule
    \end{tabular}
    \caption{Core-component ablation.}
    \label{tab:ablation_main}
\end{table}

\begin{figure}[t]
    \centering
    \includegraphics[width=\columnwidth]{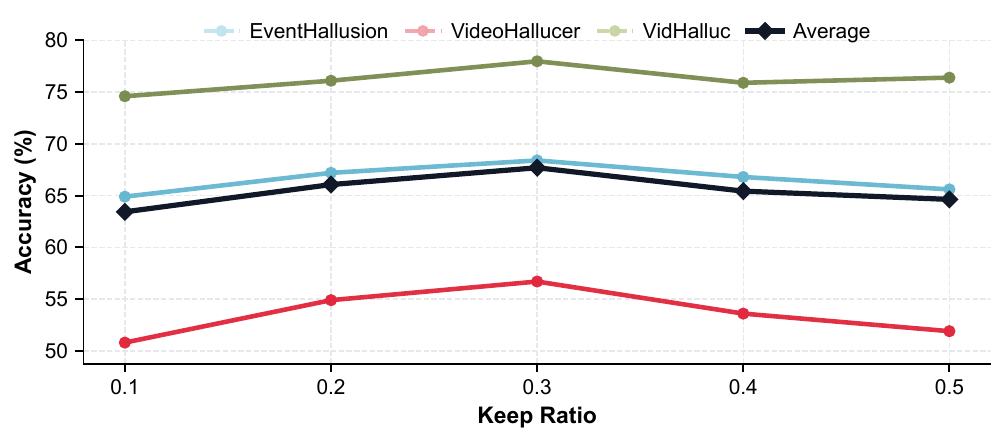}
    \caption{Sensitivity to the SEE keep ratio.}
    \label{fig:fade_keep_ratio}
\end{figure}

\noindent\textbf{SEE degradation ratio.} We study how the SEE keep ratio affects the prior-biased branch. As shown in Figure~\ref{fig:fade_keep_ratio}, performance peaks at a moderate keep ratio. A large keep ratio retains excessive discriminative evidence and weakens the contrastive signal, whereas an overly small keep ratio corrupts visual context and destabilizes suppression. Thus, SEE should moderately weaken foreground evidence rather than aggressively remove visual information.

\subsection{SEE Suppresses Prior Bias}
\begin{figure}[t]
    \centering
    \includegraphics[width=\columnwidth]{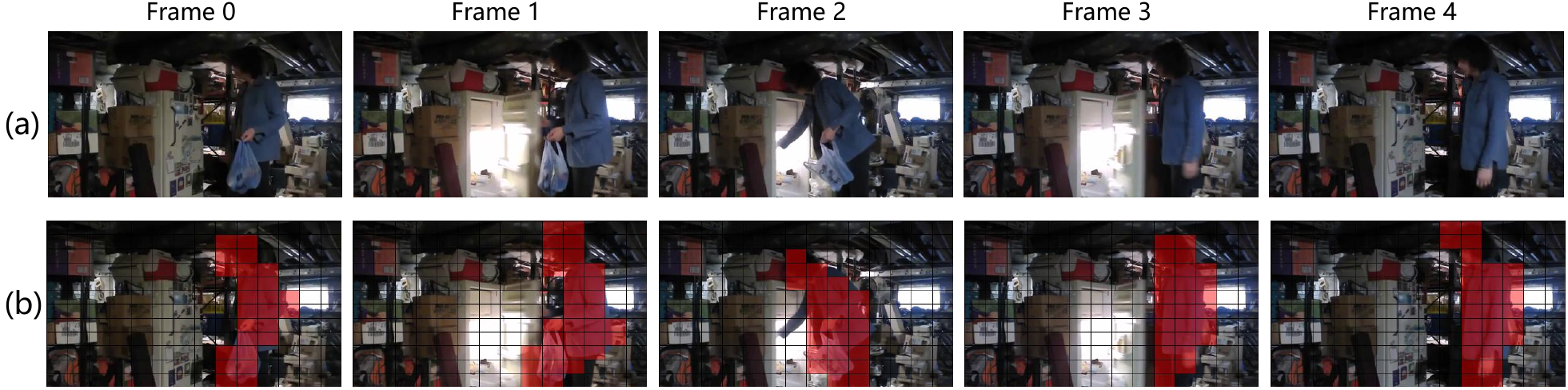}
    \par\smallskip
    \includegraphics[width=0.72\columnwidth]{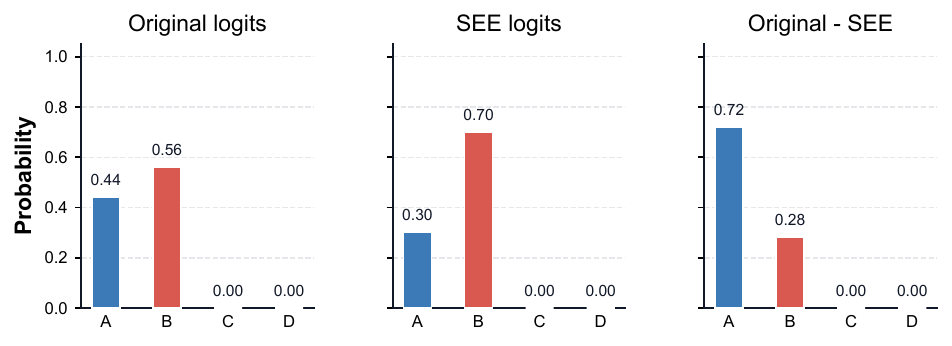}
    \caption{SEE exposes prior-driven bias through erasure and contrastive correction.}
    \label{fig:tfd_mechanism_case}
\end{figure}

Figure~\ref{fig:tfd_mechanism_case} presents a fine-grained action recognition case showing how SEE exposes prior-driven bias. Candidate A relies on localized person-object interaction, whereas the hallucination-prone candidate B is supported by coarse scene context. By erasing high-response visual tokens inside the decoder, SEE suppresses foreground evidence while retaining contextual cues, forcing the auxiliary branch to reveal the model's fallback prior. The output distributions confirm this effect: after SEE, B increases from 0.56 to 0.70, while A drops from 0.44 to 0.30. Contrastive decoding then uses this amplified prior as a negative reference, suppressing B and recovering the evidence-supported candidate A with probability 0.72. Thus, SEE is not intended to make the perturbed branch correct, but to construct a diagnostic branch that exposes hallucination bias for subsequent cancellation.

\section{Limitations}
\methodname incurs additional inference-time computation and memory from gradient-based layer selection and multi-branch forward passes. Efficient saliency estimation and branch construction are promising future directions.

\section{Conclusion}
We presented \methodname, a training-free framework separating evidence-grounded and prior-biased predictions. VFR adaptively determines where and how strongly to reallocate attention toward video evidence. SEE erases high-response video tokens per frame while retaining low-response temporal context, yielding a prior-biased branch resistant to cross-frame recovery. Across three benchmarks and multiple backbones, their joint contrastive suppression improves grounding faithfulness.

\end{document}